\documentclass[conference]{IEEEtran}
\IEEEoverridecommandlockouts

\usepackage{cite}
\usepackage{amsmath,amssymb,amsfonts}
\usepackage{caption}
\usepackage{subcaption}
\usepackage{subfloat}
\usepackage{ulem}
\usepackage{hyperref}
\usepackage{cleveref}
\usepackage{multirow}

\usepackage{graphicx}
\usepackage[export]{adjustbox}

\usepackage{algorithm}
\usepackage[noend]{algpseudocode}
\usepackage{textcomp}
\usepackage{xcolor}

\usepackage{float}
\usepackage{longtable}
\usepackage{wrapfig}
\usepackage{tikz}
\usepackage{standalone}

\usepackage[round]{natbib}

\usetikzlibrary{arrows.meta}

\usepackage{color, colortbl}
\definecolor{LightCyan}{rgb}{0.88,1,1}
\definecolor{chalkblue}{rgb}{0.671, 0.871, 0.902}
\definecolor{chalkpurple}{rgb}{0.796, 0.667,0.796}
\definecolor{chalkyellow}{rgb}{1.0, 1.0, 0.71}
\definecolor{chalkorange}{rgb}{1.0, 0.8, 0.714}
\definecolor{chalkpink}{rgb}{0.953, 0.69,0.765}

\definecolor{applegreen}{rgb}{0.55, 0.71, 0.0}
\definecolor{airforceblue}{rgb}{0.36, 0.54, 0.66}
\definecolor{amethyst}{rgb}{0.6, 0.4, 0.8}
\definecolor{antiquefuchsia}{rgb}{0.57, 0.36, 0.51}
\definecolor{aquamarine}{rgb}{0.5, 1.0, 0.83}
\definecolor{asparagus}{rgb}{0.53, 0.66, 0.42}
\definecolor{babyblue}{rgb}{0.54, 0.81, 0.94}
\definecolor{babyblueeyes}{rgb}{0.63, 0.79, 0.95}
\definecolor{babypink}{rgb}{0.96, 0.76, 0.76}
\definecolor{darkseagreen}{rgb}{0.56, 0.74, 0.56}
\definecolor{flavescent}{rgb}{0.97, 0.91, 0.56}
\definecolor{grannysmithapple}{rgb}{0.66, 0.89, 0.63}
\definecolor{pastelorange}{rgb}{1.0, 0.7, 0.28}
\definecolor{pastelmagenta}{rgb}{0.96, 0.6, 0.76}
\definecolor{richelectricblue}{rgb}{0.03, 0.57, 0.82}
\definecolor{rosevale}{rgb}{0.67, 0.31, 0.32}
\definecolor{sandstorm}{rgb}{0.93, 0.84, 0.25}
\newcommand{\R}{\mathbb{R}}

\newcommand{\post}{\text{post}}
\newcommand{\pre}{\text{pre}}

\renewcommand{\O}{\mathcal{O}}
\renewcommand{\l}{\ell}
\renewcommand{\k}{\kappa}

\def\-{\raisebox{.75pt}{-}}

\begin{document}

\title{MGiaD: Multigrid in all dimensions. \\
Efficiency and robustness by coarsening in resolution and channel dimensions
\thanks{This work is funded by the German
 Federal Ministry for Economic Affairs and Climate Action, within the project “KI Delta
Learning”, grant no.~19A19013Q. \\
contact: $\{ $betteray, rottmann, kkahl$\}$@uni-wuppertal.de
}
}
\author{Antonia van Betteray${}^{1}$, Matthias Rottmann${}^{1,2}$ and Karsten Kahl$^{1}$ \\
$1:$ University of Wuppertal, $2:$ EPFL Lausanne 
}

\maketitle

\begin{abstract}
    Current state-of-the-art deep neural networks for image classification are made up of $10$--$100$ million learnable weights and are therefore inherently prone to overfitting. The complexity of the weight count can be seen as a function of the number of channels, the spatial extent of the input and the number of layers of the network.
Due to the use of convolutional layers the scaling of weight complexity is usually linear with regards to the resolution dimensions, but remains quadratic with respect to the number of channels. Active research in recent years in terms of using multigrid inspired ideas in deep neural networks have shown that on one hand a significant number of weights can be saved by appropriate weight sharing and on the other that a hierarchical structure in the channel dimension can improve the weight complexity to linear.
In this work, we combine these multigrid ideas to introduce a joint framework of multigrid inspired architectures, that exploit multigrid structures in all relevant dimensions to achieve linear weight complexity scaling and drastically reduced weight counts. Our experiments show that this structured reduction in weight count is able to reduce overfitting and thus shows improved performance over state-of-the-art ResNet architectures on typical image classification benchmarks at lower network complexity.
\end{abstract} 

\section{Introduction}
In recent years, deep convolutional neural networks (CNNs) have proven to be among the most powerful methods for image recognition tasks \citep{krizhevsky_imagenet_2012, russakovsky_imagenet_2015, he_delving_2015}.

As a general tendency, current state-of-the-art CNN architectures for computer vision are easily comprised of $\mathcal{O}(10^7)$--$\mathcal{O}(10^8)$ learnable weights.

This enormous amount of parameters entails the risk of overfitting which can lead to poor generalization.
Thus, weight count reduction is desirable, however it may induce an undesirable bias. This trade-off is referred to as ``bias-complexity trade-off'' which constitutes a fundamental problem of machine learning, see e.g.~\citep{shalev-shwartz_understanding_2014}. 

In this work we address this problem by introducing a network architecture that achieves a more favorable bias-complexity trade-off, in terms of a parameter-accuracy trade-off, by exploiting multigrid inspired ideas. Similar to state-of-the art architectures its weight complexity scaling in resolution dimensions is linear, but significant reductions are generated by appropriate weight sharing. In addition a hierarchical structure w.r.t.\ the channel dimensions yields optimal linear scaling of weight complexity w.r.t.\ this architectural parameter. In combination we obtain an architecture whose number of weights scales only linearly in all relevant dimensions, i.e., resolution of the input and number of available channels.

In order to motivate our approach and present it in a proper context we take a brief and abstract tour into the history of neural network (NN) development. From a theoretical point of view NNs are composed of a sequence of layers, which consist of linear mappings together with biases and non-linear activation functions. The main bulk of the weights of an NN is found within the linear mappings
\[ \theta : \mathbb{R}^{m\times n\times c} \longrightarrow \mathbb{R}^{m^\prime \times n^\prime \times c^\prime}, i.e., \theta \in \mathbb{R}^{(m\cdot n\cdot c)\times (m^\prime\cdot n^\prime\cdot c^\prime)}
\] where $m, n$ and $m^\prime, n^\prime$ are the spatial dimensions and $c, c^\prime$ denote the the channel dimension of the input and output, respectively. Without any further assumptions, these linear maps are given by dense matrices, which corresponds to a fully connected layer of the NN. Such a layer then possesses $(m\cdot n\cdot c)\cdot (m^\prime\cdot n^\prime\cdot c^\prime)$ weights and becomes quickly intractable for growing $m, m^\prime$ and $n, n^\prime$~\citep{shalev-shwartz_understanding_2014}.
\begin{figure}[t]
\centering
\includegraphics[width=\linewidth]{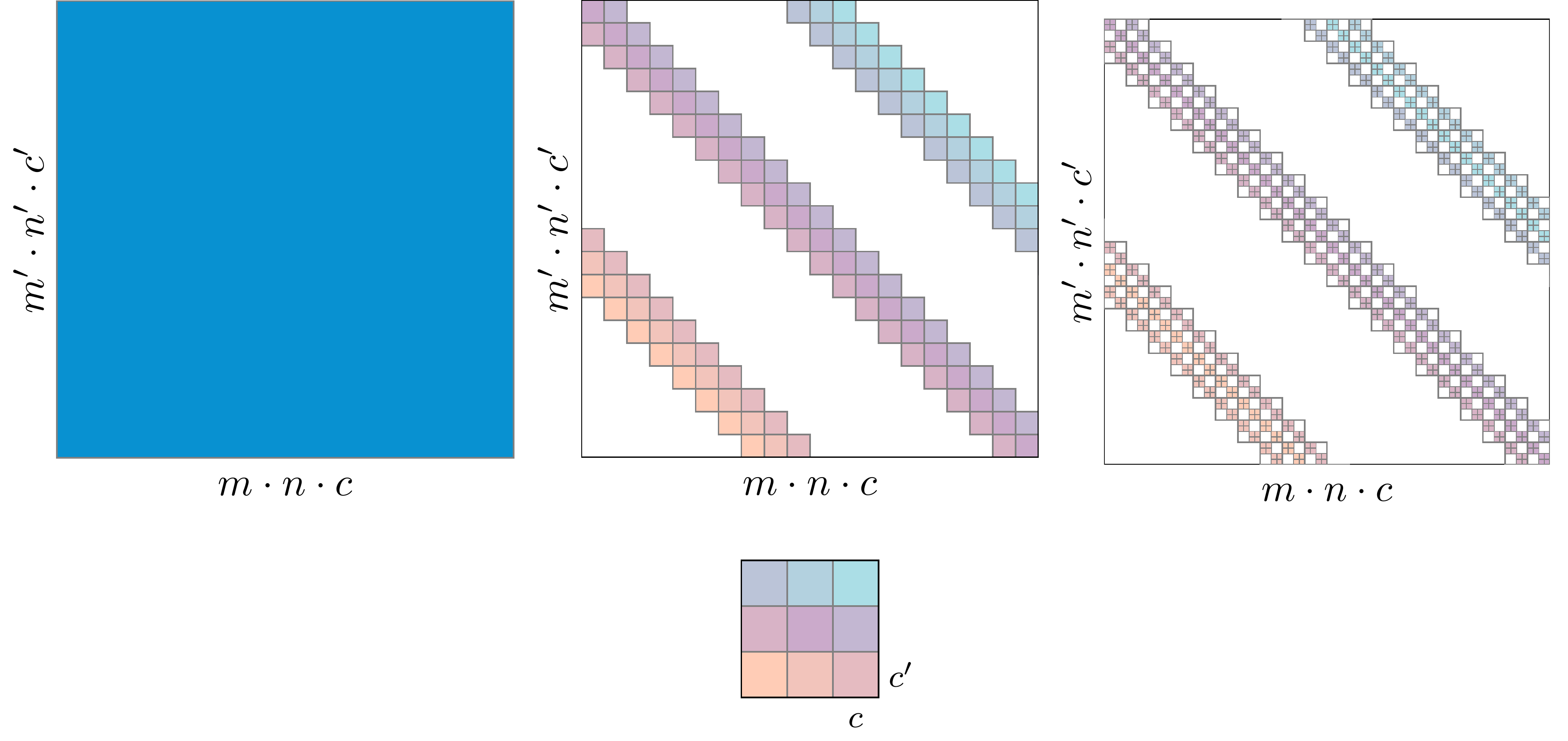}
\caption{Neural network operations in matrix representation. Fully connected layer (left), standard convolution (middle), grouped convolution (right).}\label{fig:convs_in_matrix_representation}
\end{figure}

\footnotetext{\url{https://github.com/anonymous/url/}}
{Under the assumption that learnable features are shift invariant, the first major advancement in NNs has been the introduction of convolutional neural networks~\citep{lecun_gradient-based_1998, han_learning_2015}. From the perspective of weight complexity and in particular the structure of the involved linear weight maps, these convolutional layers can be viewed as blocked, banded matrices as illustrated in~\cref{fig:convs_in_matrix_representation}. Denoting the stencil size by $s\times s^\prime$, one convolutional layer has $\mathcal{O}(s\cdot s^\prime \cdot c\cdot c^\prime)$ weights with the huge advantage that $s,s^\prime$ are fixed w.r.t.\ the resolution dimensions $m,m^\prime$ as well as $n,n^\prime.$ However, this comes at the cost of slow information exchange in these dimensions which facilitates the use of many layers and the incorporation of pooling operations to further speed up the spatial exchange.} Gating mechanisms, such as skip connections in residual networks, e.g.\ ResNets \citep{he_deep_2016, he_identity_2016}, further facilitate information flow across many layers. By gradually restoring information from feature mappings, they avoid the vanishing-gradient problem and accuracy saturation in very deep networks.

A promising approach to reduce the complexity of the network has been proposed by \citet{he_mgnet_2019}, where they make use of the inherent similarity of multigrid (MG) and residual layers, which has already been pointed out by \citet{he_deep_2016}. MG methods are hierarchical methods, typically used to solve large sparse linear systems of equations stemming from discretization of partial differential equations \citep{trottenberg_multigrid_2001}. Inspired by MG, the architecture of \citet{he_mgnet_2019}, termed MgNet, finds justification for sharing weight tensors across multiple convolutional layers in ResNet-like structures, and thus reduces the overall complexity w.r.t.\ weights by reusing layers multiple times. Still, the weight count scales quadratically w.r.t.\ the number of channels. 

Unfortunately, an assumption like shift-invariance in the spatial dimensions is amiss regarding the channel dimension and any attempt to manually sparsify its connectivity, i.e., by blocking or dropping connections, is typically met with significant performance loss. Attempts to automatically reduce CNN weight count while maintaining most of the predictive performance include pruning \citep{han_learning_2015,li_pruning_2016,he_channel_2017,wen_learning_nodate}, neural architecture search \citep{alvarez_learning_2018, gordon_morphnet_2018} as well as the development of resource-efficient architectural components \citep{howard_mobilenets_2017, sandler_mobilenetv2_2019, xie_aggregated_2017, zhang_shufflenet_2018}.

An MG perspective onto this sparsification problem is taken by~\citet{eliasof_mgic_2020}, where the artificially limited exchange of information between channels is addressed by another hierarchical structure, which resembles a multigrid $V$-cycle w.r.t.\ the channel dimensions. Using such a hierarchical construction a linear scaling of the weight count can be established.

In our work we present a weight count efficient ResNet-type architecture of MG inspired CNNs. Incorporating ideas of \citet{he_mgnet_2019} and \citet{eliasof_mgic_2020} we present a unified multigrid framework, which ultimately reduces the quadratic scaling of the weight count to linear and further reduce weights by appropriate weight sharing.

Our proposed network architecture, termed multigrid in all dimensions network (MGiaD), substantially reduces the number of weights compared to similar ResNet architectures, while maintaining performance in terms of accuracy. 

In our experiments, we compare our MGiaD architecture with multiple ResNet, MgNet and MGIC architectures on various datasets. E.g., in comparison to ResNet18 on CIFAR-10, we reduce the weight count by a factor of $10$ while sacrificing only up to $0.5$ percent points (pp) in accuracy. Even when reducing the weight count by a massive factor of $28$, the decrease in accuracy remains below 1 pp. Our implementation is available on Github \footnotemark[1].

The remainder of this article is organized as follows. First, we discuss related works in~\cref{sec:related_works}. 
In~\cref{sec:mg_and_residuallearning} we elaborate on the similarities of residual networks and MG, including the development of our MG building block. Ultimately, we present numerical experiments in~\cref{sec:experiments}.

\section{Related Works}\label{sec:related_works}
\begin{figure*}[t]
\centering
     \includegraphics[width=.8\linewidth]{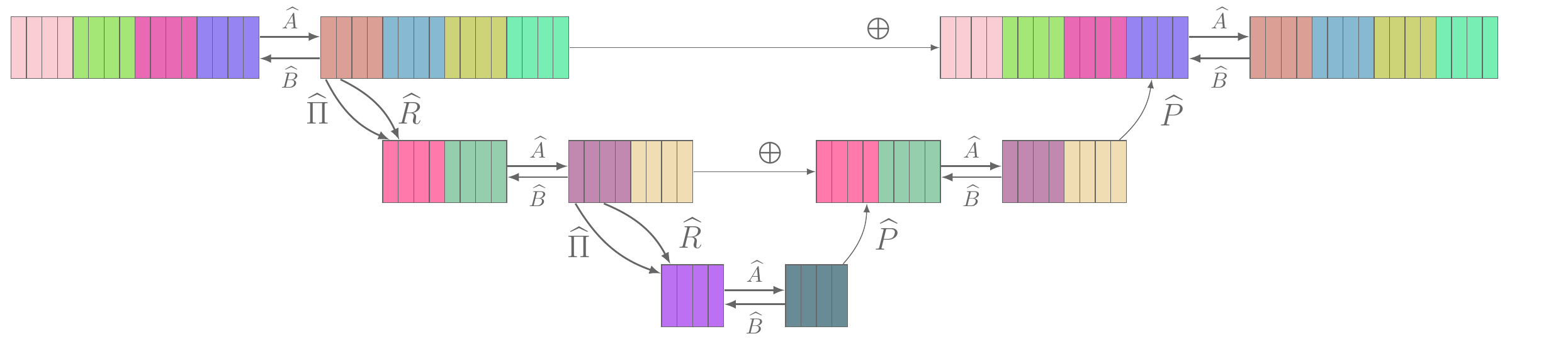}
    \caption{ $\mathtt{SiC}$-block on one resolution level, with three in channel levels and a group size of $4$. Indices are left out.}
    \label{fig:ic-block}
\end{figure*}

Work related to the MG approach we are developing in this paper can be grouped into three categories, ranging from remotely related to closely related.

\paragraph{Reducing the Number of Channels}
The reduction of weight count is oftentimes a byproduct when trying to reduce the computational footprint of a NN, e.g.\ for the sake of achieving real time capability. Though, pruning \citep{han_learning_2015, hassibi_second_nodate, li_pruning_2016, he_channel_2017} and sparsity-enhancing methods \citep{changpinyo_power_2017, han_dsd_2017} also reduce model complexity in terms of weight count while trading performance. Usually, pruning drops connections between channels after training, proving experimentally that there is redundancy in CNNs \citep{molchanov_pruning_2016}.

While the aforementioned approaches can be viewed as an automatic post-training treatment of NN, our scope is to find architectures with a reduced number of parameters pre-training that allow for a favorable rameter-accuracy trade-off when compared to post-training reductions. 

\paragraph{Modified Layers}
Another line of research is concerned with the development of convolutional layers with improved computational efficiency. Compared to the previous category, all techniques reviewed here originate from human intuition and classical methods for improving computational efficiency. One such idea 
is to use so-called depth-wise separable convolutions, that were introduced as a key feature of MobileNet architectures \citep{howard_mobilenets_2017, sandler_mobilenetv2_2019, howard_searching_2019}. Serializing the spatial dimensions, the resulting convolution kernel can be viewed as a rank one matrix of dimensions given by the kernel's spatial extent $s^2$ times the kernel's channel extent $c$. While this allows to perform convolutions with less floating point operations, the approach also reduces the number of weights in the given layer from $s^2 \cdot c$ to $s^2 + c$. 

Another approach to reduce the computational effort of convolutions consists of grouping channels \citep{krizhevsky_imagenet_2012,xie_aggregated_2017}. While the convolution usually acts within each group, the groups themselves are decoupled. In the $s^2 \times c$-matrix representation of the convolution kernel, this approach amounts to a block diagonal matrix. Due to the fact that entirely decoupling the groups hinders the distribution of information across channels, there exists approaches to circumvent this issue, such as ShuffleNet \citep{zhang_shufflenet_2018} which combines channel shuffling and grouping. Our MGiaD architecture reduces the weight count complexity without decoupling effects.

\paragraph{Multigrid-Inspired Architectures} 
In scientific computing, MG methods are known to be optimal methods for solving linear systems arising from partial differential equations (PDEs) \citep{trottenberg_multigrid_2001, treister_--fly_2011, kahl_adaptive_2018}.
These methods consist of two components that act complementary on the spectrum of the system matrix, namely the smoother and the coarse grid correction. While the former treats high frequency components, the latter treats low frequency ones. MG and deep learning have many computational components in common \citep{he_deep_2016, he_mgnet_2019}. 

The similarity of MG and CNNs also led to different architectural developments. 
\citet{ke_multigrid_2017} proposed an architecture wherein every layer is a pyramid of different scaled convolutions and every layer processes coarse and fine grid representations in parallel. \citet{he_mgnet_2019} and \citet{he_interpretive_2021} further exploited the close connection between CNNs and MG for the development of a framework called MgNet that formulates common CNN architectures as MG methods and yields a justification for sharing weight tensors across multiple layers within a given CNN architecture. MgNet utilizes MG in spatial dimensions and is capable of reducing weight counts considerably while maintaining the model's classification accuracy. 

Likewise,~\citet{eliasof_mgic_2020} achieve a reduction of the weight count by applying MG in the channels dimensions, which naturally extends grouped convolutions in an MG fashion. The resulting CNN building block is termed multigrid-in-channels (MGIC). It is built upon grouped convolutions and performs coarsening via channel pooling, thus utilizing MG in the channel dimension.
As opposed to our work, neither of the mentioned works takes a unified MG perspective onto CNNs in all dimensions.

\section{Residual Learning and Multigrid Methods}\label{sec:mg_and_residuallearning}
MG is based on two ideas: smoothing and coarse grid correction. Our focus is on image data, characterized by resolution, i.e., a grid of pixels.
Furthermore we consider the channel dimension itself as grid. 
In this section we want to explicate both smoothing and coarsening, highlighting the similarities of ResNet and MG.

\begin{figure}
\centering
\resizebox{.65\linewidth}{!}{
    \begin{subfigure}{.3\linewidth}  
        \includegraphics[height=2\linewidth]{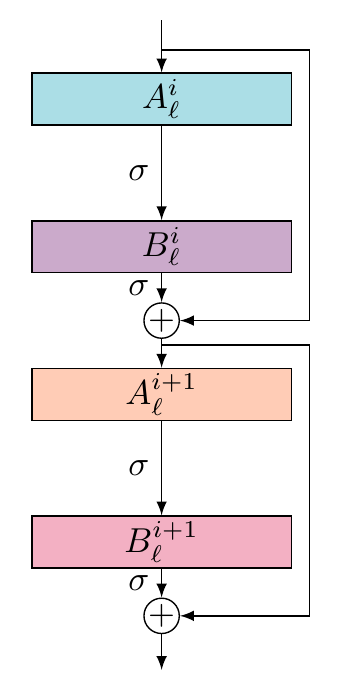}\caption{ResNet block}\label{subfig:resnet} 
    \end{subfigure}
    \begin{subfigure}{.3\linewidth}
        \includegraphics[height=2\linewidth]{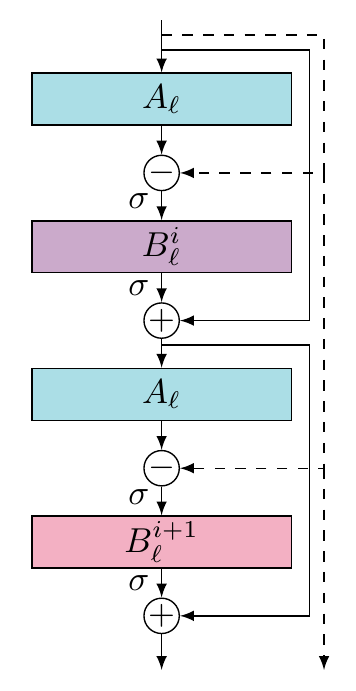}\caption{stationary}\label{subfig:mgneti} 
    \end{subfigure}
    \begin{subfigure}{.3\linewidth}
            \includegraphics[height=2\linewidth]{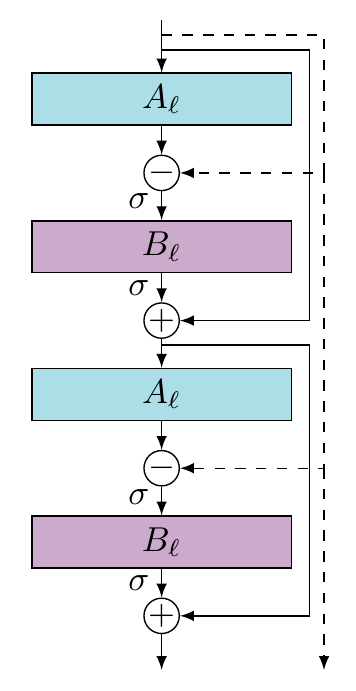}\caption{non-stationary}\label{subfig:mgnet} 
    \end{subfigure}
    }
\caption{Weight sharing in ResNet and MgNet; (\ref{subfig:resnet}) ResNet-blocks, no weight sharing; (\ref{subfig:mgneti}) MgNet-blocks, shared $A_\l$; (\ref{subfig:mgnet}) MgNet-blocks, shared layers $A_\l$ and $B_\l$.}\label{fig:resolutiondependencies}
\end{figure}

\paragraph{Revisting ResNet and MgNet}\label{subsec:resnet_and_mgnet}
 \citet{he_mgnet_2019} proposed, that the data-feature relation $A(u)=f$ can be optimized, given $A$ is learnable. The right-hand-side $f$ represents the data, $u$ belongs to the feature space and the relations between the data $f \in \R^{m \times n \times c}$ and features $u \in \R^{m \times n \times h}$ are given by 
\begin{align}
    A&:  \R^{m \times n \times h} \mapsto  \R^{m \times n \times c}, \;\;\; \text{s.t.\ } A(u) = f \label{eq:A} \\ 
    B&:  \R^{m \times n \times c} \mapsto  \R^{m \times n \times h}, \;\;\; \text{s.t.\ } u \approx B(f). \label{eq:B}
\end{align}
In that sense, we can think of $A$ as a feature-to-data map, whereas $B$ is a map that extracts features from an element of the data space.
In general $A$ and $B$ are weight matrices of convolutional layers, i.e., banded matrices. In order to explain the connection between MG and residual networks, for now we ignore the non-linear activation functions and focus of the weight matrices only. The key to understanding the iterative solution of the data-feature relation $Au = f$ is the observation that given any approximation $\widetilde{u}$ for the  solution $u$ the error $e = u - \widetilde{u}$ fulfills the residual equation
\[
Ae = A(u-\widetilde{u}) = f - A\widetilde{u} = r.
\] Hence, an approximate solution $\widetilde{e} = Br,$ defined by an appropriate feature extractor $B$, of the residual equation can be used to update the approximation $\widetilde{u} \leftarrow \widetilde{u} + \widetilde{e}$. Iterating on this idea, we obtain the general structure to solving $Au = f$ iteratively by 
\begin{equation}\label{eq:iteration}
    u = u + B^i(f - A(u)) \;\;\; \text{for} \;\;\; i = 1,2, \ldots.
\end{equation} starting with an initial guess $u = u_0$. Structurally, adding non-linear activation functions after application of any $A$ or $B^{i}$,~\eqref{eq:iteration} resembles a ResNet block as has been examined in detail by~\citet{he_mgnet_2019}. Based on the interpretation of $A$ and $B^{i}$ as data-to-feature and feature-to-data maps, respectively, it makes sense to keep $A$ fixed (i.e., use weight sharing across multiple blocks) and to either choose a different $B^{i}$ in every iteration, corresponding to a non-stationary iteration, or to keep it fixed as well turning~\eqref{eq:iteration} into a stationary iteration. The decision to keep $A$ fixed in every block can be seen as the main difference of the MgNet architecture w.r.t.\ standard ResNet. Sharing weights in $A$ and $B$ reduces the weight count, e.g., with two blocks as depicted in~\cref{fig:resolutiondependencies} we have $4 \cdot (s ^2 \times c \times h)$ without weight sharing (ResNet) and $2 \cdot (s ^2 \times c \times h)$ when sharing weights in both $A$ and $B$, i.e., considering the stationary iteration case of MgNet.

Due to the fact that we require the feature extractors $B^{i}$ to have a convolutional structure, it is clear that even if chosen optimally, this iteration is bound to converge slowly as the pseudo-inverse of $A$ is in general a dense matrix, i.e., requires a fully connected weight matrix for its representation. On the upside convolutional $B^{i}$ are cheap to apply and act locally on the data, i.e., they resolve the feature-data relation only up to a certain scale which does not encompass the whole domain. This local smoothing of features is the main observation of a MG construction, as the resulting error after a few applications of this iteration can be accurately represented on a coarser scale. 
    
\paragraph{Resolution Coarsening}\label{subsec:resoltion_coarsening}
The transfer of (residual) data to coarser scales is facilitated by mappings
\begin{equation}
    R_{\l}^{\l+1} : \R^{m_{\l} \times n_{\l} \times c_\l} \mapsto \R^{m_{\l+1} \times n_{\l+1}  \times c_{\l+1}}, 
\end{equation} resulting in a hierarchy of resolution levels for $\l =1, \ldots, L$. On each level $\l$, smoothing iterations \eqref{eq:iteration} are applied with resolution-wise mappings $A_\l$ and $B_{\l}^i$.
 
\begin{figure}[t]
 \centering
 \includegraphics[width=.75\linewidth]{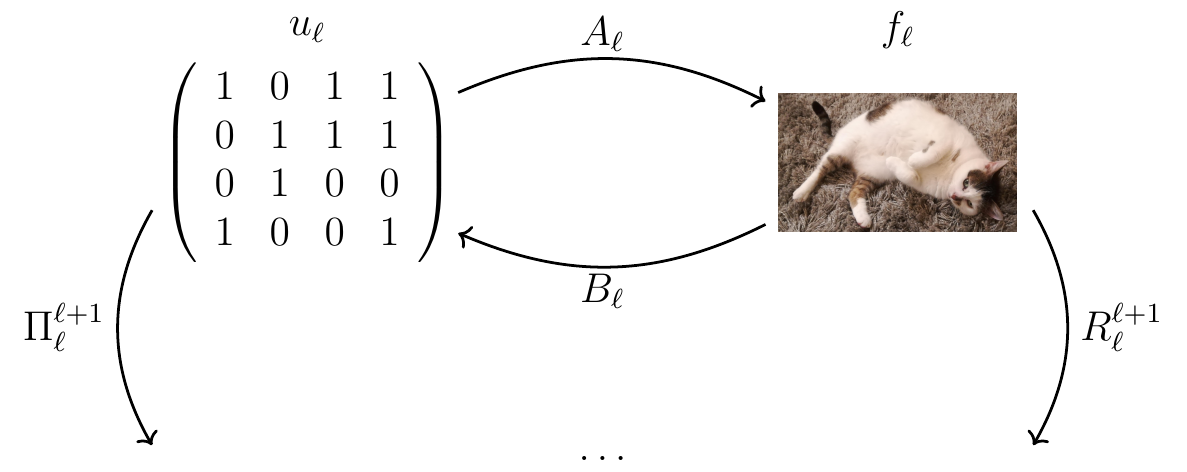}
  \caption{Data-feature relations on resolution level $\l$ followed by transfer to coarser resolution $\l+1$.}\label{fig:datafeaturerelations}
\end{figure}

The building block of CNNs that is equivalent to restrictions in MG are pooling operations, where typically resolution dimensions are reduced (by using strides $>1$) and channel dimensions are increased. Combining smoothing iterations and restrictions we obtain~\cref{alg:coarsening_multigrid}, which can be understood as the coarsening leg of a standard MG $V$-cycle \citep{trottenberg_multigrid_2001}.

\begin{algorithm}[H]
\caption{\textbackslash-MgNet($f_\l$) } \label{alg:coarsening_multigrid}
\begin{algorithmic}[1]
\State Initialization: $u_{\l} = 0$
\For{$\l = 1, \ldots, L$} 
    \For {$i = 1, \ldots, \nu$}
        \State  $u_\l = u_\l + B^{i}_\l(f_{\l}- A_\l(u_{\l}))$  
    \EndFor
    \State $u_{\l+1}=0$ 
\State $f_{\l+1} = R_{\l}^{\l+1}(f_{\l} -A_{\l} (u_{\l}))$ 
\EndFor
\end{algorithmic}
\end{algorithm}

\paragraph{Full Approximation Scheme (FAS) for Resolution Coarsening}\label{par:fullapproximationscheme}
Up to this point, we ignored the non-linearity of the overall CNN structure due to activation functions and potentially non-linear poolings and normalization operations. As usual with iterative methods for the solution of non-linear problems, the initial guess not only determines which solution is found, but also crucially influences the rate of convergence. Thus transferring the current feature approximation $u_\l$ to the coarser scale $u_{\l+1}$ can make a significant difference over choosing $u_{\l+1}=0$ as an initial guess. Consequently, in MG for the solution of non-linear problems a linear mapping 
\begin{equation}\label{eq:projection}
     \Pi_{\l}^{\l+1} : \R^{n_{\l} \times m_{\l} \times c_\l} \mapsto \R^{n_{\l+1} \times m_{\l+1} \times c_{\l+1}},
\end{equation} is introduced to initialize $u_{\l+1} = \Pi_{\l}^{\l+1}u_{\l}.$ 

Now that we start on resolution level $\l+1$ with a non-trivial initial solution, the restricted (residual) data input $f_{\l+1}$ needs to be adjusted by adding $A_{\l+1}(u_{\l +1})$. This adjustment can thus be incorporated into~\cref{alg:coarsening_multigrid} by changing 
lines $4$ and $5$ to

\begin{align}
u_{\l+1}^0 &= \Pi_\l^{\l+1}u_\l \\
f_{\l+1} &= R_l^{\l+1}(f_{\l} -A_{\l} (u_{\l})) + A_{\l+1}(u_{\l +1}).
\end{align}

Clearly, $\Pi_{\l}^{\l+1}$ corresponds to yet another pooling operation when viewed in the CNN context, but does not have any counterpart in the general ResNet architecture.
\Cref{fig:datafeaturerelations} summarizes all mappings relvant on any resolution level $\l.$

\begin{figure}
\centering
\includegraphics[width=.65\linewidth]{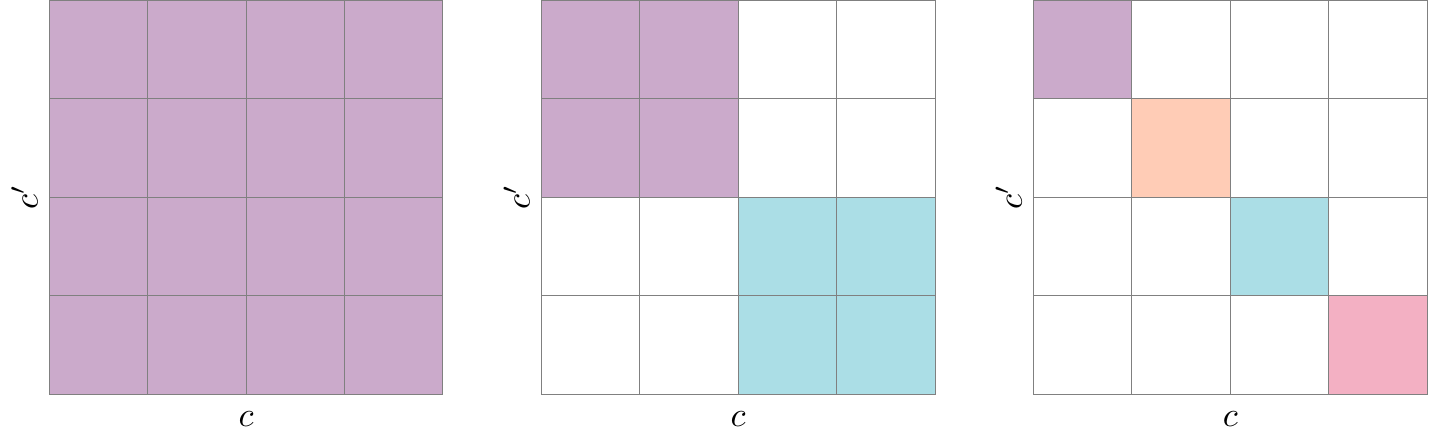}
\caption{Channel relations in convolutions: fully coupled (left), $g=1$ and $g_s= c$; grouped (middle) with $g_s=2$, s.t.\ $g = \frac{c}{2}$ and depthwise convolution $g=c$ (right).}\label{fig:coupled_convs}
\end{figure}

\paragraph{Channel Coarsening}\label{par:channel-coarsening}
While the convolutional and hierarchical structure of CNNs allows for an efficient treatment of the resolution dimensions, with a linear scaling of weights w.r.t.\ these dimensions, the situation is completely different w.r.t.\ the channel dimension. Here, typically no restriction is put on the connectivity structure, i.e., the convolutional maps are dense w.r.t.\ the coupling of input to output channels. Clearly, the full coupling of channels enables efficient exchange of information, but bears the risk of both a quadratic scaling in the number of weights and could potentially lead to a poor parameter-accuracy trade-off due to redundancies. Assuming that the number of channels cannot be reduced, a reduction in the number of weights is only possible by addressing their connectivity. Unfortunately, one cannot profit from an invariance assumption that allowed the introduction of convolutional connections in the resolution dimensions. Thus any reduction in the channel connectivity is of an ad-hoc nature. The most straight-forward way of limiting the connectivity is that of grouping channels and limiting exchange to happen only within each group (cf.~\cref{fig:coupled_convs}). Denoting the group size by $g_s$ such a strategy reduces the number of weights from $O(s^2 \cdot c \cdot h)$ to $\O(s^2 \cdot \frac{c\cdot h}{g})$ \citep{krizhevsky_imagenet_2012}. 

While replacing $A_\l$ and $B_\l$ (or only $A_\l$ or $B_\l$) by grouped convolutions with group sizes $g_s < c$ reduces the weight count significantly, the lack of interaction between the channels also results in a decrease of performance in terms of accuracy, see ~\cref{fig:perf_vs_complexity}. In order to facilitate efficient interaction of channels \citet{eliasof_mgic_2020} introduced a grouped restriction, 

\begin{equation}\label{eq:in-channel-restriction}
    \widehat{R}^{\k+1}_{\l,\k} : \R^{m_{\l} \times n_{\l} \times c_{\l, \k} } \mapsto  \R^{m_{\l} \times n_{\l} \times c_{\l, \k + 1}}
\end{equation}
with $c_{\l,\k + 1} = (\frac{c_{\l, \k}}{2})$ that reduces the number of channels by a factor of $2$ as depicted in~\cref{fig:ic-block}.

In correspondence to a MG $V$-cycle (cf.~\citep{trottenberg_multigrid_2001}) another grouped mapping
\begin{equation}\label{eq:in-channel-prolongation}
    \widehat{P}^{\k}_{\l,\k+1}: \R^{m_{\l} \times n_{\l} \times c_{\l, \k + 1} } \mapsto  \R^{m_{\l} \times n_{\l} \times c_{\l, \k}}
\end{equation}
is introduced to refine the number of channels, interpolating coarse level features $\widehat{u}_{\l, \k+1}$ to the fine level, starting from the coarsest which uses a dense CNN-block, e.g.\ a ResNet-block \citep{eliasof_mgic_2020}. Similar to MG such a coarse-level update on level $\k$ is given by

\begin{equation}\label{eq:coarsegrid-correction}
    \widehat{u}_{\l, \k} = \widehat{u}_{\l, \k} + \widehat{P}_{\l,\k+1}^{\k} (\widehat{u}_{\l, \k+1})
\end{equation}
Using this MG strategy for the channel dimensions allows for a significant reduction in the number of weights without sacrificing much accuracy~\citep{eliasof_mgic_2020}.

\paragraph{Multigrid in all Dimensions: MGiaD}
In order to obtain an architecture that ultimately scales linearly in the number of weights w.r.t.\ all problem dimensions (resolution and channels), combining the ideas of~\cref{sec:mg_and_residuallearning} $a)$ and $c)$, we introduce a (MG) smoothing in channels (\texttt{SiC}), that uses \eqref{eq:in-channel-restriction} and \eqref{eq:in-channel-prolongation} to build an in-channel hierarchy, that incorporates smoothing iterations \eqref{eq:iteration} with shared weights w.r.t.\ the in-channel level $\k$.

\begin{figure}[t]
\centering
\includegraphics[width=.65\linewidth]{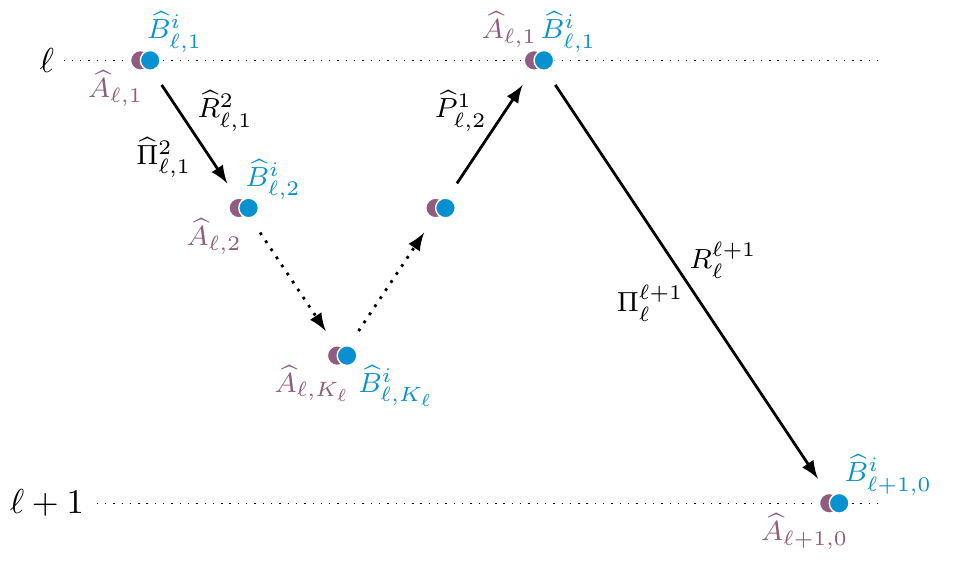}
\caption{Structure of \texttt{SiC} on resolution levels.}\label{fig:MGiaD_structure}
\end{figure}

To be more precise we replace the maps $A_{\l}$ and $B_{\l}^{i}$ in~\cref{alg:coarsening_multigrid} by an in-channel $V$-cycle in the following way. The convolutions $A_{\l}$ and $B_{\l}^{i}$, which are fully connected w.r.t.\ the channel dimensions are replaced by grouped convolutions $\widehat{A}_{\l,1}$ and $\widehat{B}^{i}_{\l,1}$, respectively. In addition we introduce grouped convolutions $\widehat{A}_{\l,\k}, \widehat{B}^{i}_{\l,\k}$ for $\k = 1,2,\ldots, K_{\l}$ as well as restrictions $\widehat{R}_{\l,\k}^{\k+1}$ and interpolations $\widehat{P}_{\l,\k+1}^{\k}$ for $\k = 1,2,\ldots, K_\l-1$. In here the number of levels $K_{\l}$ is chosen such that $\widehat{A}_{\l,K_\l}$ and $\widehat{B}_{\l,K_\l}$ are again fully connected w.r.t.\ the channel dimension.
The overall structure of the resulting method is sketched in~\cref{fig:MGiaD_structure}.

On each channel level $\k$ the grouped convolutions $\widehat{A}_{\l,\k}$ and $\widehat{B}^{i}_{\l,\k}$ are arranged as in~\eqref{eq:iteration}, i.e., as an in-channel smoothing iteration. Analogous to the introduction of an FAS-type restriction $\Pi_{\l}^{\l+1}$ of the current feature map in MgNet, we introduce an in-channel FAS restriction map
\begin{equation}\label{eq:in-channel-projection}
    \widehat{\Pi}^{\k +1}_{\l,\k} : \R^{m_{\l} \times n_{\l} \times c_{\l, \k+1} } \mapsto  \R^{m_{\l} \times n_{\l} \times c_{\l, \k}}.
\end{equation} Clearly, both $\widehat{R}_{\l,\k}^{\k+1}$ and $\widehat{\Pi}_{\l,\k}^{\k+1}$ have to be grouped mappings as well in order to end up with a linear scaling of the number of weights w.r.t.\ the channel dimension size.

The final algorithm of the MG smoothing in-channel algorithm (\texttt{SiC}) is summarized in~\cref{alg:in-channel-MG}.
 
\begin{algorithm}[H]
\caption{Multigrid smoothing in channels $\mathtt{SiC}(f_{\k}, \, u_{\k})$}\label{alg:in-channel-MG}

\begin{algorithmic}[1]

\For {$i = 1, \ldots, \eta_{\pre}$}
        \State  $u_{\k} = u_{\k} + B_{\k}^{i}(f_{\k}- A_{\k} u_{\k})$  \Comment{pre-smoothing}
\EndFor
\If{$\k \neq K$} 
\State $u_{\k+1} = \widehat{\Pi}_\k^{\k+1}(u_{\k})$
\State $f_{\k+1} = \widehat{R}_\k^{\k+1}(f_{\kappa} - A_{\k}(u_{\k})) + A_{\k+1}(u_{\k+1})$
\State $\widehat{u}_{\k+1} = \mathtt{SiC}(f_{\k+1}, u_{\k+1})$
\State $u_\k = u_\k + \widehat{P}_{\k+1}^\k (\widehat{u}_{\k+1})$
\EndIf
\For {$i = 1, \ldots, \eta_{\post}$} \Comment{post-smoothing}
        \State  $u_{\k} = u_{\k} + B_{\k}^{i}(f_{\k}- A_{\k}( u_{\k}))$   
\EndFor
\end{algorithmic}

\end{algorithm}

The dependency of the number of weights is reduced from quadratic to linear scaling w.r.t.\ the channel dimension when compared to a fully connected structure as in ResNet or MgNet. 
Replacing the smoothing iteration in MgNet by in-channel-MG-blocks finally yields a MG-like architecture that achieves linear scaling of the number of weights in all dimensions. The resulting method termed multigrid in all dimensions (MGiaD) is given in~\cref{alg:mgiad}.

\begin{algorithm}[H]
\caption{Multigrid in all dimensions $\mathtt{MGiaD}(f_\l)$}\label{alg:mgiad}

\begin{algorithmic}[1]
\State Initialization $u_1 = 0$
\For{$\l = 1, \ldots, L-1$} 
    \State $u_\l = \mathtt{SiC}(f_\l, u_{\l})$
    \State $u_{\l+1} = \Pi_\l^{\l+1}(u_\l)$
    \State $f_{\l+1} = R_l^{\l+1}(f_{\l} -A_{\l} (u_{\l})) + A_{\l+1}(u_{\l +1})$
\EndFor
\end{algorithmic}
\end{algorithm}

\section{Experimental Setup \& Evaluation}\label{sec:experiments}
We evaluate our approach on improving the parameter-accuracy trade-off for classifaction tasks on different popular datasets such as CIFAR-10, CIFAR-100 ~\citep{krizhevsky_learning_2009} and FashionMNIST~\citep{xiao_fashion-mnist_2017}.
We report the number of weights, train and test accuracy with standard deviation (std). Since the standard deviation for the train accuracy is $\text{std $\leq 0.01$}$, it is omitted in the following.

\paragraph{Training Setup}
The implementation of our models is based on Pytorch~\citep{paszke_pytorch_2019}.
Unless otherwise stated, we train the models with batch-size $128$ for $400$ epochs with an SGD-optimizer, a momentum set to $0.9$ and a weight decay of $10^{-4}$. In accordance to ResNet~\citep{he_deep_2016} we use batch normalization followed by a ReLU activation function after every convolutional layer. The initial learning rate is set to $0.05$ and we use a cosine-annealing learning rate scheduler~\citep{loshchilov_sgdr_2017} to adapt the learning rate during training.

\paragraph{Baselines Models and Architecture} 
We compare our MGiaD approach to ResNet18, ResNet20 and the corresponding MgNet architectures. The ResNet18 architecture is composed of $4$ resolution levels with $[64, 128, 256, 512]$ channels and $2$ ResNet-blocks on each resolution level. According to \citet{he_mgnet_2019} we reduce the number of channels on the last resolution level from $512$ to $256$. ResNet20 on the other hand, which is a ResNet version specifically designed for classification on CIFAR-10, is made up of $3$ resolution levels \`{a} $3$ blocks and $[16, 32, 64]$ channels. The parameters of the MgNet architectures are chosen accordingly in either situation. We include tests of the MgNet architecture with sharing $A$ and/or $B$ convolutions, which we refer to by MgNet$^{\rm A}$ when sharing only $A$ convolutions and MgNet$^{\rm A,B}$ when sharing both $A$ and $B$. 

Based on the findings in~\cref{tab:exp_baselineResNetMgNet} 
and by~\citet{he_interpretive_2021}, sharing $A$ and $B$ operators leads to a significant reduction in weight count when compared to the corresponding ResNet architectures at a miniscule performance loss of $0.2$ pp. Thus, we opt to default to sharing both $A$ and $B$ convolutions when it comes to showing results for the MGiaD architecture. In all tests we chose learnable depthwise pooling operators $\Pi$ and $R$ in both MgNet and MGiaD. 

\begin{figure}[t]
\begin{center}
    \includegraphics[width=0.99\linewidth]{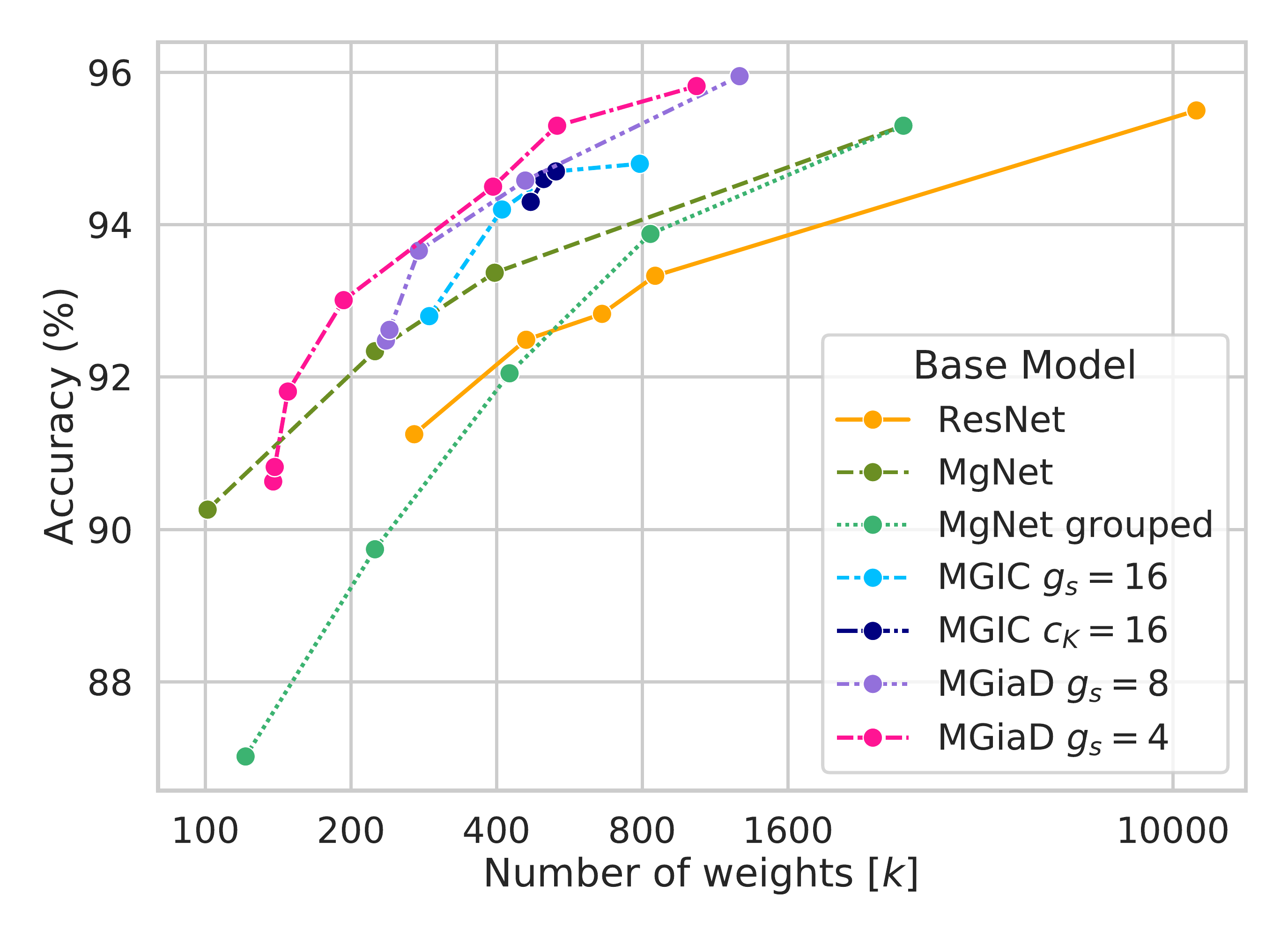}
     \caption{Parameter-accuracy trade-off on CIFAR-10 for different ResNet, MgNet, MGIC~\citep{eliasof_mgic_2020} and MGiaD models.
     For MGiaD $g_s=4$ and $g_s =8$, the number of channels on the coarsest grid varies, i.e., $c_K \in \{4, 8, 16, 32, 64\}$.}
    \label{fig:perf_vs_complexity}
\end{center}
\end{figure}

As an additional comparison, we also include tests where channel grouping in the MgNet architectures is used as a mean of additional sparsification. We demonstrate that without using a hierarchical structure supporting information exchange between all channels, we obtain significant performance losses, see~\cref{fig:perf_vs_complexity} (MgNet grouped).
 
\begin{table}[t]
    \centering
    \caption{Test and training accuracy and weight count of ResNet20 ($3$ resolution layers), ResNet18 ($4$ resolution layers) and corresponding MgNet models trained on CIFAR-10. In each section, the model with highest accuracy is highlighted in bold face. The model with highest overall accuracy is highlighted with purple background.}\label{tab:exp_baselineResNetMgNet}
    \scalebox{0.8}{
    \begin{tabular}{l | r | c | c }
        \hline
            \multirow{2}{*}{\textbf{ Model}} & \multirow{2}{*}{$\#$ \textbf{weights}} & \multicolumn{2}{c}{\textbf{accuracy} $\pm$ \textbf{std}}  \\ \cline{3-4}
             & &  test & train \\
      
         \hline 
           ResNet20 &   $270k$ &  $ \mathbf{92.44 }$ ($ 0.16 $) & $ 93.54 $   \\
           MgNet$^{\rm A}$  & $198k$ &   $ 91.47 $ ($ 0.28 $) & $ 91.41 $  \\
           MgNet$^{\rm A,B}$ &  $101k$ &   $ 90.58 $ ($ 0.30 $) & $ 88.92 $   \\
         \hline 
 \rowcolor{chalkpurple!30}         ResNet18 &  $11,174k$ & $ \mathbf{95.58} $ ($ 1.56 $) & $ 97.71 $  \\
         MgNet$^{\rm A}$ &  $4,\!115k$ &     $ 95.38 $ ($ 0.03 $) & $ 97.48 $    \\
         MgNet$^{\rm A,B}$ & $2,\!751k$ &  $ 95.28 $ ($ 0.04 $) & $ 97.26 $   \\
         \hline
             \end{tabular}}
\end{table}

\paragraph{Evaluation on CIFAR-10}\label{par:CIFAR-10}
In order to identify the important parameters for improving the parameter-accuracy trade-off, we examine a large-scale parameter study on CIFAR-10, which contains $60k$ color images of size $32 \times 32$ pixels in $10$ classes.

We start our study with a review of the considered baselines in~\cref{tab:exp_baselineResNetMgNet}. As expected we see that sharing convolutions within the ResNet blocks significantly reduces the number of weights of the models at minor decreases in performance. Note, that the loss in performance is larger for the baseline model of lower complexity (ResNet20). 
We test the use of grouped convolutions in the $A$ and $B$ part of the MgNet blocks in the larger MgNet models. According to this test we can rule out the possibility of a trivial sparsification w.r.t.\ the channel dimension in this architecture. While grouped convolutions lead to a significant reduction in weight count, the loss of information sharing in channel dimension is clearly seen in the quite substantial performance loss, see~\cref{fig:perf_vs_complexity} (MgNet grouped). More detailed results can be found in the appendix.

With the stage set by these baseline tests we can turn our attention to our MGiaD architecture. In~\cref{tab:groupsize_and_coarsestgrid} we report results with varying group sizes and size of coarsest level convolutions within the channel MG subcycle. The results clearly indicate that the size of the fully connected coarsest level of the MG subcycle has a significant influence on the performance of the resulting network, while the group size of the grouped convolutions $\widehat{A}$ and $\widehat{B}$ has a much smaller impact. At $g_s=4$ and $c_{K} = 64$ we obtain an architecture with a weight count $30$ times smaller compared to ResNet18 at a cost of only $1$ pp of accuracy. Even compared to the slim MgNet$^{\rm A,B}$, this architecture yields comparable performance at a $7\times$ reduction in weight count. Compared to the smaller ResNet20 model with only $270k$ weights we find for $g_{s}=8$ and $c_{K} = 32$ an MGiaD model with a similar number of weights, but a substantially improved performance and for $g_{s} =4$ and $c_{K} = 16$ a model with roughly half the number of weights and comparable accuracy.

 \begin{table}[t]
    \centering
    \caption{Influence of numbers of channels on the coarsest level $c_K$ and group size $g_s$ on accuracy and weight count for the MGiaD architecture, compared to ResNet18, ResNet20 and corresponding MgNet$^{\rm A,B}$ models on CIFAR-10.
    }\label{tab:groupsize_and_coarsestgrid}
    \scalebox{0.8}{
    \begin{tabular}{l | r | r | r|  r | c | c }
    \hline
    \multirow{2}{*}{\textbf{ Model}} & 
    \multirow{2}{*}{$c_K$}& 
    \multirow{2}{*}{$g_s$}&
    \multirow{2}{*}{$\lambda$}&
    \multirow{2}{*}{$\#$\textbf{weights}}  & 
    \multicolumn{2}{c}{\textbf{accuracy} $\pm$ \textbf{std}} \\ \cline{6-7}
        & & & & & test & train  \\
   \hline
    ResNet20 &- &-  & - & $270k$ & $ \mathbf{92.44}$ ($ 0.16 $) & $ 93.54 $  \\
    MgNet$^{\rm A,B}$ & -&- & - & $101k$ &  $ 91.26 $ ($ 0.09 $) & $ 90.88 $  \\
    \hline
    \hline
    ResNet18 &- & - & - &  $11,\!174k$ &  $ \mathbf{95.58} $ ($ 1.56 $) & $ 97.71 $ \\
    MgNet$^{\rm A,B}$ &  - & - & - & $2,751k$ &    $ 95.28 $ ($ 0.04 $) & $  97.26 $  \\
    \hline
    MGiaD &   $4$ & $4$ & $1$ & $138k$ & $90.63$ ($ 0.16$)  & $ 88.72 $  \\
    MGiaD &   $8$ & $4$ & $1$ &  $139k$ &  $90.82$ ($ 0.14$) & $ 88.93 $  \\
    MGiaD &  $16$ & $4$ & $1$ &  $148k$ &  $91.81$ ($ 0.35$) & $ 90.41 $  \\
    MGiaD &  $32$ & $4$ & $1$ & $193k$ &  $93.01$ ($ 0.22$) &$ 93.13 $   \\
    MGiaD &  $64$ & $4$ & $1$ & $393k$ &  $\mathbf{94.62}$ ($ 0.10$) &$ 95.83 $   \\
    \hline
    MGiaD &  $ 8$ & $8$& $1$ & $236k$ &   $92.48$ ($0.37$) & $ 92.46 $    \\
    MGiaD &  $16$ & $8$& $1$ & $240k$ &  $92.62$ ($0.16$) & $ 92.69 $ \\
    MGiaD &  $32$ & $8$& $1$ & $276k$ &  $93.66$ ($ 0.25$)& $ 94.29 $   \\
    MGiaD &  $64$ & $8$ & $1$ & $458k$ &  $\mathbf{94.58}$ ($ 0.25$)  & $ 96.11 $  \\
    \hline
    MGiaD &  $16$ & $16$ & $1$ & $424k$ &  $94.13$ ($0.15$) & $ 95.29 $   \\
    MGiaD &  $32$ & $16$ & $1$ & $441k$ &  $94.31$ ($ 0.14$)& $ 95.70 $   \\
    MGiaD &  $64$ & $16$ & $1$ & $586k$ &  $\mathbf{94.72}$ ($ 0.35$) & $ 96.44 $  \\
    \hline
    MGiaD &  $32$ & $32$ & $1$ & $773k$ &  $94.97$ ($0.37$) &$ 96.65 $   \\
    MGiaD &  $64$ & $32$ & $1$ & $845k$ &  $\mathbf{95.20}$ ($0.20$) & $ 96.81 $   \\
    \hline
    MGiaD &  $64$ & $64$ & $1$ & $1,\!361k$ &  $\mathbf{95.47}$ ($0.07$) & $ 97.23 $   \\
    \hline
     MGiaD &  $64$ & $4$ & $3$ & $1,\!020k$ & $  95.64$ ($0.09$) &  $ 97.21 $  \\
\rowcolor{chalkpurple!30}     MGiaD &  $64$ & $8$ & $3$ & $1,\!269k$ & $\mathbf{95.95}$ ($0.12$) & $ 97.44 $   \\
     \hline
    \end{tabular}
    }
\end{table}
 To utilize the freed up capacity in terms of weights and pursue high accuracy, we now introduce a channel scaling parameter $\lambda$. We use this parameter to scale the initial number of channels fixing $c_{K} = 64$ and $g_{s} = 4$ and $g_{s} =8$. The larger number of overall channels leads to a deeper hierarchy, more parameters and improved performance~\cref{tab:groupsize_and_coarsestgrid}. A detailed discussion can be found in the appendix.\\ 
 Another venue of performance improvement is the (re)-use of $\widehat{A}$ and $\widehat{B}$ in the channel MG subcycle in a fashion akin to post-smoothing in MG~\citep{trottenberg_multigrid_2001}, where such a process is known to speed up the time and work to solution of the problem. Correspondingly, we add multiple post-smoothing steps $\eta_\post$ to the in channel hierarchy. As the weight tensors are shared on one channel level, only weights for new batch normalizations are added to the total. For most setups, multiple post-smoothing steps do not improve the accuracy, see also the appendix.

To summarize our extensive tests for CIFAR-10 we compiled all results in terms of weight count vs.\ test accuracy in~\cref{fig:perf_vs_complexity}, where we also include results for the MGIC approach described by~\citet{eliasof_mgic_2020}. What we can see is that we obtain very good results using $g_s = 4$ and varying values of $c_{K}$ and $\lambda$. However, there is some diminishing returns starting to set in around $500k$ weights and the best result we were able to achieve uses $g_{s}=8$. 
\paragraph{Evaluation on FashionMNIST}\label{par:fashionMNIST}
FashionMNIST contains
$70k$ $32 \times 32$ greyscale images belonging to $10$ classes. The initial learning rate is chosen as $0.05$ and multiplied by $0.1$ every $25$ epochs.
The results are presented in~\cref{tab:fashionmnist}. In all our experiments we observe strong overfitting.
However, MGiaD still improves the parameter-accuracy trade-off. In particular, in comparison to ResNet18, MGiaD for $c_K=g_s=64$ achieves the same accuracy with 8$\times$ less weights. Additionally, the weight count for MGiaD can be reduced by another factor of $3$ while sacrificing only $0.1$ pp in performance. 
In order to reduce overfitting, we decreased the number of resolution levels from $4$ to $3$ for the experiments with $c_K=16$, see~\cref{tab:fashionmnist_small}. We compare the resulting architecture with a ResNet20, also consisting of $3$ resolution levels. We observe indeed a mild increase in performance along side with a reduction in weight count of a factor of $2.7$.
\begin{table}[t] 
    \centering
    \caption{Influence of number of channels $c_K$ and group size $g_s$ on FashionMNIST. ResNet18 and corresponding MgNet for comparison. 
    }\label{tab:fashionmnist}
    \scalebox{0.8}{
    \begin{tabular}{l | r | r | r |  c | c}
        \hline
        \multirow{2}{*}{\textbf{Model}} & \multirow{2}{*}{$c_K$}  & \multirow{2}{*}{$g_s$} & \multirow{2}{*}{$\#$\textbf{weights}} & \multicolumn{2}{c}{\textbf{accuracy} $\pm$ \textbf{std}} \\\cline{5-6}
        & & & &  test & train \\ 
         \hline        
         ResNet18 & - & - &$11,\!003k$&  $ \mathbf{93.84} $ ($ 0.17 $) & $  100$\\
         MgNet & - & - &$2,\!747k$ & $ \mathbf{93.84} $ ($ 0.16 $) & $  100$  \\
        \hline
        MGiaD & $64$ & $4$ & $389k$&  $ 93.45 $ ($ 0.12 $) & $  100$ \\
        MGiaD & $64$ & $8$& $454k$ &  $ 93.50 $ ($ 0.30 $)   & $  100$\\
        MGiaD & $64$ & $64$ & $1,\!357k$&  $ \mathbf{93.84} $ ($ 0.10 $) & $  100$ \\
        \hline
         MGiaD & $32$ & $4$ & $189k$& $ 93.32 $ ($ 0.06 $) & $  100$\\
\rowcolor{chalkpurple!30} MGiaD & $32$ & $32$ & $437k $& $\mathbf{93.71} $ ($ 0.49 $)  & $  100$ \\
        \hline
         MGiaD & $16$ & $4$ & $420k$ & $\mathbf{93.40}$ ($ 0.21 $)& $  100$\\
         MGiaD & $16$ & $16$ & $144k$ & $ 93.28 $ ($ 0.08 $)  & $  100$   \\
        \hline
    \end{tabular}}
\end{table}
\begin{table}[t]
    \centering
    \caption{Influence of number of fully coupled channels $c_K$ and the group size $g_s$ on FashionMNIST, compared to ResNet20 and corresponding MgNet.}
    \label{tab:fashionmnist_small}
    \scalebox{0.8}{
    \begin{tabular}{l | r | r | r |  c | c}
        \hline
        \multirow{2}{*}{\textbf{Model}} & \multirow{2}{*}{$c_K$}  & \multirow{2}{*}{$g_s$} & \multirow{2}{*}{$\#$\textbf{weights}} & \multicolumn{2}{c}{\textbf{accuracy} $\pm$ \textbf{std}} \\ \cline{5-6}
        & & & &  test & train \\ 
         \hline        
         ResNet20 & - & - &$270k$&  $ 93.02 $ ($ 0.31 $)& $  100$\\
         MgNet & - & - &$101k$ & $\mathbf{93.29} $ ($ 0.15 $) & $  100$  \\
        \hline
        MGiaD & $16$ & $4$ & $28k$& $ 92.85 $ ($ 0.06 $) &  $ 97.07 $ \\
 \rowcolor{chalkpurple!30}  MGiaD & $16$ & $8$ & $37k$ & $ \mathbf{93.35 }$ ($ 0.09 $)& $ 97.63 $ \\
        MGiaD & $16$ & $16$ & $55k$ & $ 93.29 $ ($ 0.15 $) & $ 98.48 $ \\
        \hline
    \end{tabular}}
\end{table}    

\paragraph{Evaluation on CIFAR-100}\label{CIFAR-100}
CIFAR-100 contains $100$ classes with $600$ images each (same specs as CIFAR10). Due to the fact that we observed for CIFAR-10 and FashionMNIST, that a high number of fully connected channels has a significant influence on the accuracy, we opt to choose $g_K=64$. In~\cref{tab:CIFAR-100} we study the influence of the group size $g_s$ and the number of channels on the accuracy when varying the channel multiplier $\lambda$. Similar to CIFAR-10 we observe that a higher weight count in general leads to better accuracy. But, compared to CIFAR-10 where the group size had a minor impact, for CIFAR-100 a big group size is essential for a high accuracy. For instance, a model with $g_s=8$ and $481k$ weights achieves an accuracy of almost $70$ pp, whereas its counterpart with $g_s=64$ achieves an accuracy of over $72$ pp with a way higher number of weights, i.e., more than $1,\!300k$. In accordance to our CIFAR-10 results, the channel scaling parameter $\lambda$ successfully utilizes the freed-up capacity. Multiplying the number of channels by $3$ leads to the best overall accuracy. The resulting model has half the number of weights of ResNet18, but improves upon its accuracy by $0.4$ pp.
 
\begin{table}[t]
    \centering
    \caption{Influence of the total number of channels w.r.t.\ the resolution levels, scaled by $\lambda$ on CIFAR-100. The number of fully coupled channels is $c_K = 64$ and the group size is $g_s \in \{8, 64\}$. MGiaD models are compared to ResNet18 and corresponding MgNet.
}\label{tab:CIFAR-100}
    \scalebox{0.8}{
    \begin{tabular}{l | r | r  | r | c | c }
       \hline
        \multirow{2}{*}{\textbf{Model}} & 
        \multirow{2}{*}{$\lambda$} &
        \multirow{2}{*}{$g_s$} &
        \multirow{2}{*}{$\#$\textbf{weights}}& 
        \multicolumn{2}{c}{\textbf{accuracy} $\pm$ \textbf{std}} \\ \cline{5-6}
       
         & & & & test & train \\
         \hline
        ResNet18 & - & - & $11,\!220k$& $ \mathbf{75.42} $ ($ 0.13 $) & $ 99.98 $ \\
        MgNet &- & - &$2,\!774k$&  $ 74.42 $ ($ 0.28 $) & $ 99.98 $ \\
       \hline
       MGiaD & $1$ &  $8$  & $481k$   &   $69.91$  ($0.37$) & $ 99.25 $  \\
       MGiaD & $1$ &   $64$ & $1,\!384k$ &  $\mathbf{72.53}$ ($0.45$) & $ 99.97 $  \\
       \hline
       MGiaD & $2$ &   $8$  & $745k$  &  $71.48$ ($0.48$) & $ 99.91 $ \\
       MGiaD & $2$ &   $64$  & $ 3,067k$ &   $\mathbf{75.12}$  ($0.71$) & $ 99.98 $\\
       \hline
       MGiaD & $3$ &  $8$  & $1,\!338k$ & $ 72.75 $ ($ 0.62 $) & $ 99.97 $  \\
\rowcolor{chalkpurple!30}  MGiaD & 3 &  64  & $4,822k$& $\mathbf{75.85}$ ($0.14$) & $ 99.98 $  \\
       \hline
    \end{tabular}}
\end{table}

\section{Conclusion}\label{sec:conclusion}
In this work we introduced a neural network architecture that utilizes the concept of multigrid in spatial and channel dimensions. Our experiments suggest that, although the reduction in weight count introduces an additional architectural bias, this bias in most cases seems not to affect the overall performance of the network. This holds in particular when the network is over-parameterized. In problems requiring only limited capacity, e.g.\ CIFAR-10, the proposed architecture indeed yields improved results while substantially reducing the weight count. Our approach offers another way to account for over-parameterization of deep neural networks and achieves an improved scaling behavior w.r.t.\ the depth and width hyperparameters. In future work, it will be of interest to study how the smoother $B$ can be replaced by a polynomial in $A$ to additionally reduce the weight count.
\section{Main Assumptions and Limitations}
Our implementation is based on pytorch \citep{paszke_pytorch_2019}. Since grouped convolutions have lesser computational operations than fully coupled convolutions, we expect a speed-up in training time as well. However, this is not achieved since the underlying implementation for grouped convolution is not efficient on GPUs
\footnote{\url{https://github.com/pytorch/pytorch/issues/70954}} \footnote{\url{https://github.com/pytorch/pytorch/issues/73764}}\footnote{\url{https://github.com/pytorch/pytorch/issues/18631}}.

\section{Appendix}\label{sec:appendix}
\section{Supplemental Experiments}
The following supplemental experiments disseminate some aspects of our multigrid in all dimensions architecture in more detail. These include experiments with grouped convolutions in MgNet and investigations of the influence of a channel scaling parameter $\lambda$ and post-smoothing in MGiaD, on both CIFAR-10 and CIFAR-100. 

The experimental setup is identical to the main paper.
\subsection{Evaluation on CIFAR-10}
\paragraph{Grouped Convolutions in MgNet}
To justify the use of a hierarchical structure in the channel dimension, we discuss replacing fully coupled convolutions by grouped convolutions in the ResNet-type architecture MgNet. The reduced number of connections between the grouped blocks reduces the weight count and thus the channel interaction. In~\cref{tab:mgnet_groupedconvolutions} the influence of the group size $g_s$ on the number of weights and the accuracy is reported. Compared to ResNet18 and MgNet, both with fully coupled convolutions, smaller group sizes $g_s$ in $A$ and $B$ in MgNet convolutions lead to a drastic reduction of the weight count, but also of the accuracy. A model with convolutions of a rather large group size $g_s=32$ has $425k$ parameters but only achieves an accuracy of $92$ pp. The highest accuracy of MgNet with grouped $A$ and $B$ achieved with a $g_s=64$ with $831k$ weights. These results suggest a naive sparsification of the channel dimension without a hierarchical structure is not easily achieved.

\begin{table}[h]
\centering
    \caption{Results of MgNet with different group sizes $g_s$ in $A$ and $B$ convolutions in a $4$ resolution layer MgNet as well as ResNet18 as a reference point on CIFAR-10. The group size for a fully coupled convolution equals to the number of channels $c$. In each table section the model with highest overall accuracy is highlighted with purple background.
    }\label{tab:mgnet_groupedconvolutions}
    \scalebox{0.8}{
    \begin{tabular}{l | c | r | c | c  }
    \hline
     \multirow{2}{*}{\textbf{ Model}} & \multirow{2}{*}{$ \text{ \textbf{$g_s$}}$} & \multirow{2}{*}{$ \boldsymbol{\#}$ \textbf{weights}} & \multicolumn{2}{c}{\textbf{accuracy} $\pm$ \textbf{std}}  \\ \cline{4-5}
             & &  & test & train \\
 \hline
\rowcolor{chalkpurple!30}          ResNet18 & $c$ &  $11,\!174k$ &  $\mathbf{95.58} $ \;($ 1.56 $) & $ 97.71 $ \\
        MgNet$^{\rm A,B}$&  $c$ & $2,\!751k$ &  $ 95.28 $ \;($ 0.04 $) & $ 97.26 $  \\
          \hline 
        MgNet$^{\rm A,B}$ & $64$ & $ 831k$  &   $ \mathbf{93.88} $ \;($ 0.15 $)  & $ 95.50 $ \\
        MgNet$^{\rm A,B}$ & $32$ & $ 425k$  &   $ 92.05 $\; ($ 0.28 $)  & $ 92.86 $ \\
        MgNet$^{\rm A,B}$ & $16$ & $ 223k$  &   $ 89.74 $\; ($ 0.59 $)  & $ 88.08 $ \\
        MgNet$^{\rm A,B}$ & $8$ & $ 121k$   &   $ 87.02 $\; ($ 0.76 $)  & $ 82.32 $ \\
        \hline
    \end{tabular}}
\end{table}

\paragraph{Channel Scaling in MGiaD} 
In previous studies we showed, that a group size of $4$ or $8$ with $c_K=64$ fully connected channels on the coarsest in channel level reduces the accuracy while maintaining accuracy. To utilize the freed-up capacity in terms of weight we introduce a channel scaling parameter $\lambda$, which is multiplied to the initial number of channels. We report results for $\lambda \in \{1, 2, 3 \}$ in~\cref{tab:cifar10_channelscaling}. Multiplying the number of channels naturally increases the the number of weights linearly, which has beneficial effect on the accuracy. For a group size of $8$ and $\lambda=3$ the resulting model achieves almost $96$ pp in accuracy, with $1,\!269k$ parameters, which is half the weight count of MgNet and a factor $8$ of the ResNet18 parameters. This shows that, as long as the overall cost w.r.t.\ the channel dimension is linear, as in MGiaD, trading channel connectivity for channel dimension is beneficial.

\begin{table}
    \centering
    \caption{Number of initial channels multiplied by $\lambda \in \{ 1,2,3 \}$, with $g_K=64$ and $g_s=4$ and $g_s=8$ in MGiaD. ResNet and MgNet results for comparison.}
   
    \label{tab:cifar10_channelscaling}
\scalebox{0.8}{
    \begin{tabular}{l | c | c | r | c | c }
    \hline
    \multirow{2}{*}{\textbf{ Model}} & 
    \multirow{2}{*}{$g_s$} & 
    \multirow{2}{*}{ $\lambda$} &    
    \multirow{2}{*}{$\#$\textbf{weights}} &
    \multicolumn{2}{c}{\textbf{accuracy} $\pm$ \textbf{std}}  \\ \cline{5-6}
      & & & & test & train \\
       \hline 
        ResNet18 & $c$ &  -& $11,\!174k$ &  $\mathbf{95.58} $ \;($ 1.56 $) & $ 97.71 $ \\
        MgNet$^{\rm A,B}$&  $c$ &  -&$2,\!751k$ &  $ 95.28 $ \;($ 0.04 $) & $ 97.26 $  \\
         
    \hline
    MGiaD &$4$ & $1$ &  $393k$ &  $94.62 \; (0.10)$ & $95.83$\\ 
    MGiaD & $4$ & $2$ & $533k$ &  $95.02 \; (0.21)$ & $96.32$ \\
    MGiaD & $4$& $3$ & $1,\!020k$ &   $ \mathbf{95.64} \; (0.09)$ & $97.21$ \\
   \hline
    MGiaD &  $8$ & $1$ &$458k$ &  $94.58$ \; ($ 0.25$)  & $ 96.11 $\\
    MGiaD & $8$& $2$ & $533k$ &   $95.02$ \;($ 0.21$) & $96.32$  \\
\rowcolor{chalkpurple!30}  MGiaD & $8$& $3$ & $1,\!269k$& $\mathbf{95.95}$\; ($0.12$) & $ 97.44 $ \\
    \hline
    \end{tabular}}
\end{table}

\paragraph{Post-smoothing in MGiaD}
From the multigrid perspective taken in the main document it is obvious to ask for a variation of the number of smoothing iterations performed on each level of the channel hierarchy. Due to weight sharing of the convolutions an increase in smoothing iteration amount to a marginal increase in parameter. Thus as a last study, we report results of a slightly more complex MGiaD model with $g_s =8$ and $g_s=4$ and $\lambda = 3$ for varying number of post-smoothings $\eta_\post$ in~\cref{tab:cifar10_gs4_gs8_postsmoothing}. Consistently to observations made in~\cref{tab:cifar10_channelscaling} and table $2$ in the main paper, a bigger group size improves the accuracy, while the effect of number of post-smoothing iterations is ambiguous.

\begin{table}

    \centering
    \caption{Influence of post-smoothing $\eta_\post$ in MGiaD with channel scale of $\lambda=3$, $c_K=64$ and $g_s \in \{4, 8\}$ on CIFAR-10.}\label{tab:cifar10_gs4_gs8_postsmoothing}
    \scalebox{0.8}{
    \begin{tabular}{l | c | c |r | c | c }
    \hline
    \multirow{2}{*}{\textbf{Model}} &  
    \multirow{2}{*}{\textbf{$g_s$}} &
    \multirow{2}{*}{\textbf{$\eta_\post$}} & 
    \multirow{2}{*}{$\#$\textbf{weights}} &
    \multicolumn{2}{c}{\textbf{accuracy} $\pm$ \textbf{std}}  \\ \cline{5-6}
          & & &  &  test & train \\
     \hline
     MGiaD & $4$ & $1$ & $1,\!020k$ & $ 95.64 $ ($0.09$)  & $ 97.21 $ \\
     MGiaD & $4$ & $2$ & $1,\!035k$ & $\textbf{95.83}$ ($0.13$) & $ 97.25 $ \\
     MGiaD & $4 $& $3$ & $1,\!051k$ & $ 95.53$ ($0.17$)  & $ 97.32 $ \\
     \hline
 \rowcolor{chalkpurple!30}     MGiaD & $8$&$1$ & $1,\!269k$& $\mathbf{95.95}$ ($0.12$) & $ 97.44 $ \\
     MGiaD & $8$&  $2$ & $ 1,\!276k$&  $ 95.70 $ ($ 0.26 $) & $ 97.35 $ \\
     MGiaD & $8$&  $3$ & $1,\!300k$&  $ 95.90$ ($0.23$) & $ 97.31 $\\
        \hline
    \end{tabular}}
\end{table}

\subsection{Evaluation on CIFAR-100}
Similar to CIFAR-10 we study the influence of the number of channels scaled by $\lambda$ and the number of post-smoothings w.r.t.\ the channel levels. The results are reported in~\cref{tab:CIFAR-100_post}. Consistently to observations made for CIFAR-10,~\cref{tab:cifar10_channelscaling} and~\cref{tab:cifar10_gs4_gs8_postsmoothing}, increasing the channel dimension is beneficial for the accuarcy, while the impact of post-smoothing has a negligible effect.
\begin{table}[h]
    \centering
    \caption{Influence of number of channels w.r.t.\ resolution levels, scaled by $\lambda$ and post-smoothing $\eta_{\post}$ for MGiaD on CIFAR-100. The number of fully coupled channels is $c_K = 64$ and the group size is $g_s \in \{8, 64\}$. ResNet18 and corresponding MgNet results for comparison.
    }\label{tab:CIFAR-100_post}
    \scalebox{0.8}{
    \begin{tabular}{l | c | c | r | r | c | c }
       \hline
        \multirow{2}{*}{\textbf{Model}} & 
        \multirow{2}{*}{$\lambda$} &
        \multirow{2}{*}{$\eta_\post$} &
        \multirow{2}{*}{$g_s$} &
        \multirow{2}{*}{$\#$\textbf{weights}}& 
        \multicolumn{2}{c}{\textbf{accuracy} $\pm$ \textbf{std}} \\ \cline{6-7}
         & & & & & test & train \\
         \hline
        
         ResNet18 & - &- & -& $11,\!220k$& $\mathbf{ 75.42} $ \;($ 0.13 $) & $ 99.98 $\\
         MgNet & - & -& -&$2,\!774k$&  $ 74.42 $\; ($ 0.28 $) & $ 99.98 $  \\
       \hline 
       \hline
        MGiaD & $1$ &  $1$ & $8$  & $481k$   &   $69.91$ \; ($0.37$) & $ 99.25 $ \\
        MGiaD & $1$ &  $1$ & $64$ & $1,\!384k$ &  $\mathbf{72.53}$ \;($0.45$) & $ 99.97 $ \\
        \hline
        MGiaD & $1$ &  $3$ & $8$  &  $490k$ & $70.19 $\;  ($0.10$)  & $ 99.35 $ \\
        MGiaD & $1$ &  $3$ & $64$ & $1,\!393k$  &  $\mathbf{72.74}$\; ($0.73$) &  $ 99.96 $  \\
       \hline
       \hline
       MGiaD & 3 & 1 & 8  & $1,\!338k$ & $ 72.75 $ \; ($ 0.62 $) & $ 99.97 $  \\
\rowcolor{chalkpurple!30}  MGiaD & 3 &  1& 64  & $4,\!822k$& $\mathbf{75.85}$\; ($0.14$) & $ 99.98 $  \\
      \hline 
       MGiaD & $3$ & $3$ & $8$  & $1,\!338k$ & $72.75$\; ($ 0.38$) & $ 99.96 $  \\
       MGiaD & $3$ &  $3$& $64$  & $4,\!853k$&  $\mathbf{75.40}$\; ($0.35$) & $ 99.97 $  \\
       \hline
    \end{tabular}}
\end{table}

\setlength{\itemindent}{-\leftmargin}

\bibliography{literatur}

\end{document}